\DocumentMetadata{pdfstandard=A-2b,pdfversion=1.7,lang=en-US}
\documentclass[nohead,nofoot]{asmeconf}
\usepackage{booktabs}
\usepackage{tabularx}
\usepackage{xcolor}
\usepackage{placeins}
\usepackage[nameinlink,noabbrev]{cleveref}

\Crefname{equation}{Equation}{Equations}
\Crefname{figure}{Figure}{Figures}
\Crefname{table}{Table}{Tables}

\newcommand{\method}{OpenQlaw-v2}

\hypersetup{
  pdftitle={OpenQlaw-v2: A Unified Agentic AI Assistant for Analysis of 2D Quantum Materials},
  pdfauthor={Sankalp Pandey and coauthors},
  pdfsubject={Preprint},
  pdfkeywords={agentic AI, multimodal language models, 2D materials, microscopy, tool use}
}

\begin{document}

\title{\NoCaseChange{OpenQlaw-v2}: A Unified Agentic AI Assistant for Analysis of 2D Quantum Materials}

\SetAuthors{%
  Sankalp Pandey\affil{1},
  Thanh-Dat Truong\affil{1},
  Tim Faltermeier\affil{2}\affil{3}\affil{5},
  Nicholas Borys\affil{3}\affil{5},
  Hugh Churchill\affil{4}\affil{5},
  Khoa Luu\affil{1}\affil{5}
}
\SetAffiliation{1}{Quantum AI Lab, University of Arkansas, Fayetteville, AR}
\SetAffiliation{2}{Department of Physics, Montana State University, Bozeman, MT}
\SetAffiliation{3}{Department of Physics and Astronomy, University of Utah, Salt Lake City, UT}
\SetAffiliation{4}{Department of Physics, University of Arkansas, Fayetteville, AR}
\SetAffiliation{5}{MonArk NSF Quantum Foundry}

\maketitle


\begin{abstract}
Interactive inspection of two-dimensional (2D) quantum materials requires visual characterization and reasoning over successive requests. The existing OpenQlaw workflow uses a general-purpose agent to coordinate a separate material-analysis expert, requiring two model deployments. In this work, we introduce OpenQlaw-v2, a unified agentic artificial intelligence (AI) assistant that extends a physics-aware multimodal model from a material-analysis expert to the central reasoning agent. We adapt its language component using tool-use conversations generated from executable inspection tasks. 
The same checkpoint operates in two modes. Task reasoning is performed with the adapter enabled, while image analysis uses the original model with the adapter disabled.
Image-specific records retain observations and calibration across requests, supporting deterministic counting, ranking, calibrated bounding-box measurement and visual annotation. We compare OpenQlaw-v2 with the original two-model workflow and an unadapted domain expert. The results show that adaptation improves tool coordination while preserving the original characterization outputs. The unified approach reduces serving requirements and supports multi-turn inspection, although the separate orchestrator completes full workflows more reliably. These findings demonstrate the potential of adapting a domain expert into an inspection agent for material selection and fabrication planning.
\end{abstract}

\section{Introduction}
Identifying atomically thin flakes with precise layer thicknesses and sizes is a bottleneck in the manufacturing of next-generation quantum devices \cite{nguyen2026quacod,nguyen2025qmoe,
nguyen2024qclusformer,holliday2025quadro,
nguyen2024quantum,nguyen2024hierarchical,
nguyen2025diffusion,holliday2026advanced,
holliday2024hybrid} from two-dimensional (2D) materials.
Atomic force microscopy (AFM) provides local surface measurements, but manually finding and inspecting individual flakes makes this process slow and labor-intensive~\cite{saito2019}.
Instead, optical microscopy captures many flakes across a substrate in a single image and allows for rapid inspection of layer characteristics~\cite{blake2007making}.
Automating their identification can greatly reduce the manual effort required to select materials for fabrication.

\begin{figure*}[!t]
    \centering
    \includegraphics[width=0.8\textwidth]{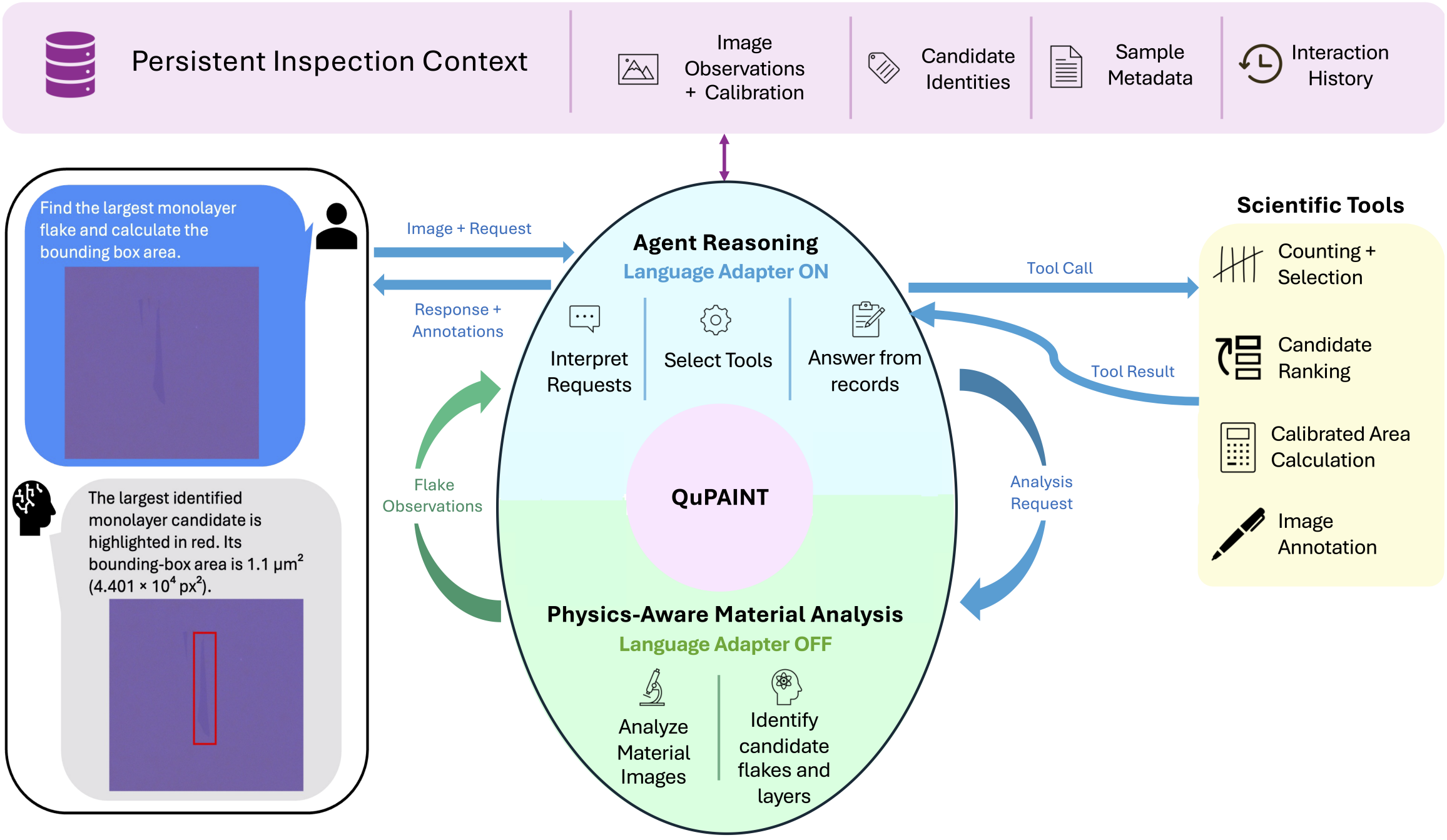}
    \caption{\textbf{Overview of OpenQlaw-v2.} The approach employs a single domain expert model (QuPAINT) operating in two modes. Agent reasoning is used for request interpretation, tool orchestration and response generation. Physics-aware material analysis for flake identification and layer classification. Deterministic scientific tools provide quantitative analysis and annotation, while persistent context maintains information across interactions.}
    \label{fig:architecture}
\end{figure*}

\subsection{Flake Characterization to Agentic Automation}
Traditional deep learning approaches, such as object detection and segmentation, have been proposed to automate this perception task~\cite{masubuchi2020deep,nguyen2024two,uslu2025maskterial,uslu2024open,nguyen2025varphiadaptphysicsinformedadaptationlearning,pandey2025cliff,zhu2020deformable,carion2020end,redmon2016you,ronneberger2015u,kirillov2023segment,ravi2024sam}.
However, reliable identification must account for how layer thickness, substrate interference, and imaging conditions affect flake appearance.
Flakes from different classes may look similar, whereas flakes within the same class may have different appearances across images~\cite{nguyen2026qupaintphysicsawareinstructiontuning,nguyen2025varphiadaptphysicsinformedadaptationlearning}.
QuPAINT~\cite{nguyen2026qupaintphysicsawareinstructiontuning} addresses these challenges by incorporating material-specific physical knowledge into a multimodal model for quantum material characterization.

Tool-using language models connect reasoning to external operations~\cite{yao2023react,schick2023toolformer,truong2026agentic}, including interpreters for numerical computation~\cite{gao2023pal}.
In microscopy, Experiment Automation Agents (EAA) use vision-language models to coordinate instrument operations, including focusing, feature search and data acquisition~\cite{du2026eaa}.
OpenQlaw~\cite{pandey2026openqlawagenticaiassistant} focuses on interactive analysis of 2D materials, using a general-purpose agent to coordinate QuPAINT and deterministic tools for counting, selection, ranking, measurement and annotation.
Image-specific records retain the observations and calibration needed for successive requests.

\subsection{Domain Expert as the Central Agent}
While OpenQlaw supports interactive material inspection, its architecture requires separately hosting the general-purpose agent and the material-analysis expert.
Separately hosting these models increases serving requirements and requires transferring observations between them.
This motivates using QuPAINT itself as the central agent for both task reasoning and material analysis.
However, QuPAINT is trained to produce structured material characterizations rather than select tools and execute inspection requests.
The model must therefore learn to interpret researchers' requests, invoke the appropriate tools, and use their results across successive interactions.
At the same time, adapting the model for tool use should preserve its original material-characterization behavior.
This requires supporting agent decisions and image analysis within the same checkpoint.

\subsection{Contributions}
In this work, we introduce OpenQlaw-v2, a unified agentic AI assistant that positions QuPAINT as the central agent for interactive inspection of 2D quantum materials.
Our contributions are threefold.
First, we adapt the domain expert to interpret requests, select scientific tools and answer from their results, replacing the separately hosted general-purpose orchestrator.
Second, we use a low-rank language adapter for tool-use instruction tuning and switch between adapter-enabled agent reasoning and adapter-disabled characterization within the same loaded checkpoint.
The existing scientific tools and image-specific inspection records are retained from OpenQlaw.
Third, we evaluate the unified approach against the two-model workflow and an unadapted domain expert, examining task completion, tool-use accuracy, consistency across interactions and computational requirements.
\section{The Proposed OpenQlaw-v2 Approach}
\label{sec:method}
OpenQlaw-v2 extends the original OpenQlaw framework by positioning QuPAINT to handle both task reasoning and material analysis.
In the original framework~\cite{pandey2026openqlawagenticaiassistant}, a general-purpose language model interprets requests and coordinates the domain expert (QuPAINT) and the tools.
We replace this separate orchestrator by adapting QuPAINT to interpret requests, select tools, and generate responses from their results.
The observation records, scientific tools, and inspection memory are retained from OpenQlaw.
\Cref{fig:architecture} illustrates our proposed approach.

\subsection{Domain Expert as the Central Agent}
\label{sec:sharedmodel}
Originally, QuPAINT is explicitly trained for physics-aware material characterization, where an optical microscopy image and an instruction produce an analysis of the candidate flakes~\cite{nguyen2026qupaintphysicsawareinstructiontuning}.
However, acting as the central agent requires the model to also decide which operation is needed to address a researcher's request.
For example, comparing candidate sizes requires ranking the observations, while calculating physical dimensions requires a measurement tool and an image scale.

To produce this behavior, we attach a low-rank adapter to QuPAINT's language component~\cite{hu2021lora}.
The original model parameters remain frozen during training.
We formulate inspection and analysis as a sequence of agent decisions over the current request and available observations.
At decision step $t$, the inspection state is
\begin{equation}
s_t = (I_t, R_t, C_t, H_t),
\label{eq:inspection-state}
\end{equation}
where $I_t$ contains the images, $R_t$ contains their observation records and candidate identities, $C_t$ contains the available image scales, and $H_t$ contains the interaction history.
The runtime maintains this state and supplies the relevant image identifiers, records, scales and history as agent context.
Image pixels are passed to the model when the image-analysis tool is executed.
The state in \Cref{eq:inspection-state} connects successive requests to the corresponding images and candidates.

Given the current user request $u$, the adapted domain expert produces
\begin{equation}
a_t = \pi_{\theta,\Delta}(u,s_t),
\label{eq:agent-action}
\end{equation}
where $\theta$ denotes the frozen QuPAINT parameters and $\Delta$ denotes the trainable agent adapter.
The action in \Cref{eq:agent-action} is a tool call with its name and arguments or a final response to the researcher.
The request $u$ remains fixed across the decision steps needed to answer it.
The inspection state and scientific tools are retained from OpenQlaw. The adapted QuPAINT model assumes responsibility for selecting the actions.

The same loaded checkpoint supports both agent reasoning and material analysis.
For an adapted language projection $W$, the effective weights are
\begin{equation}
W_{\mathrm{eff}} =
\begin{cases}
W + \dfrac{\alpha}{r}BA, & \text{agent reasoning}, \\[4pt]
W, & \text{material analysis},
\end{cases}
\label{eq:roles}
\end{equation}
where $A$ and $B$ are trainable low-rank factors, $r$ is the adapter rank, and $\alpha$ controls the update scale.
Together, these factors across the selected language projections form $\Delta$.
As shown in \Cref{eq:roles}, the adapter is active by default during agent execution.
When QuPAINT produces an image-analysis tool call, the runtime temporarily disables the adapter and runs the same loaded model using its original characterization prompt.
The resulting observations are stored in the inspection records.
The runtime then restores the adapter and returns the observations to the agent for subsequent operations.
The two modes execute sequentially without loading a second checkpoint.
This allows QuPAINT to perform both roles without separately hosting a general-purpose reasoning model.

\subsection{Tool-Use Instruction Tuning}
\label{sec:adaptation}
We train the agent adapter using conversations constructed from executable inspection tasks.
Each training example contains a user request and the available inspection context, followed by a tool call or an answer supported by the tool result.
The scientific tools execute the requested operations to provide the numerical values and observation records used in these conversations.

The training tasks cover counting, candidate selection, ranking, annotation, and calibrated bounding-box measurement.
They also include requests with missing information, such as an unavailable image scale or absent layer labels.
These examples teach the model when to execute an operation and when to ask for clarification.
Multi-turn conversations further teach the model to use previously identified candidates and measurements when answering follow-up requests.
The conversations also include requests that begin without an observation record.
In these examples, the agent first invokes the image-analysis tool and then uses the returned observations to select subsequent operations.
Synthetic analysis responses provide this supervision without using microscopy images for agent adaptation.

We train the adapter to produce the actions defined in \Cref{eq:agent-action} from the corresponding requests and inspection states.
Let $h_i$ denote the serialized training context containing the request, available state, tool definitions, and preceding interaction.
Let $y_i$ denote the target assistant response, containing either a tool call or a final answer.
We optimize the adapter using
\begin{equation}
\mathcal{L}_{\mathrm{agent}}(\Delta)
=
-\frac{1}{N}\sum_{i=1}^{N}\frac{1}{n_i}
\sum_{j=1}^{n_i}
\log p_{\theta,\Delta}
\left(y_{i,j}\mid h_i,y_{i,<j}\right),
\label{eq:training}
\end{equation}
where $N$ is the number of training examples and $n_i$ is the number of supervised tokens in the target response, including its terminator.
The objective in \Cref{eq:training} supervises the designated assistant response.
User requests, observation records, and tool results provide context and receive no direct training loss.
The original parameters $\theta$ remain frozen, and only the adapter parameters $\Delta$ are updated.
The adaptation therefore targets the agent's tool-use behavior while leaving the original characterization weights unchanged.

\subsection{Scientific Tools and Multi-Turn Inspection}
\label{sec:execution}
\label{sec:tools}
OpenQlaw-v2 uses the scientific tools and image-specific records established in OpenQlaw.
Each record retains the source image identity, candidate identities, and observations returned by material analysis.
Counting, selection, ranking, measurement, and annotation operate over these identified records.
A proposal record contains the available candidate enumeration, while a conclusion record contains the subset selected by the material analysis.
The requested operation must use the corresponding record and observation scope.

For each tool call, the runtime validates the proposed operation and its arguments before execution.
For a valid call, the tool result and updated inspection state are
\begin{equation}
\begin{aligned}
o_t &= \mathcal{T}(a_t,s_t), \\
s_{t+1} &= \mathcal{U}(s_t,a_t,o_t),
\end{aligned}
\label{eq:state-update}
\end{equation}
where $\mathcal{T}$ executes the selected operation and $\mathcal{U}$ updates the retained records and interaction history.
For image analysis, $\mathcal{T}$ invokes the original characterization mode in \Cref{eq:roles}.
For scientific operations, it executes the corresponding deterministic tool.
The updated state in \Cref{eq:state-update} is returned to the agent to select another operation or generate a final response through \Cref{eq:agent-action}.
Invalid calls produce an error and are handled according to the configured validation policy.
The interaction ends when the agent produces a final response or reaches the execution limit.

Physical measurements additionally require a scale supplied for the corresponding image.
For a bounding box with width $w$ and height $h$ in pixels, and an isotropic scale $\rho$ in micrometers per pixel, the measurement tool computes
\begin{equation}
A_{\mathrm{box}} = wh\rho^2 \quad [\mu\mathrm{m}^2].
\label{eq:area}
\end{equation}
Here, $\rho$ is the pixel scale stored in $C_t$.
The operation in \Cref{eq:area} is retained from OpenQlaw and measures bounding-box area rather than segmented flake area.
The agent receives the actual tool result and generates the response. 

Inspection records and calibration remain available across successive requests within the session.
Researchers can therefore refine their selection, request an annotation, or repeat a measurement without requiring another image analysis when the necessary observations are already available.
The resulting workflow follows \Cref{fig:architecture}, with QuPAINT controlling the scientific operations previously coordinated by the separate orchestrator.
\section{Experimental Setup}
\label{sec:setup}

\subsection{Data and Inspection Tasks}
We evaluate task reasoning on 60 episodes from six synthetic scenes not used for adaptation, with shared task templates.
The tasks cover counting, selection, ranking, annotation, physical-area measurement and requests with missing information.
We also evaluate inspection on 24 optical microscopy images. Twelve of these are with monolayer graphite. Six are with non-monolayer crystallites, and six are empty images.
Positive images are used for counting, ranking, annotation, calibrated measurement and repeated measurement requests.
The negative examples include MoS$_2$, WS$_2$, WSe$_2$ and hBN.
All evaluation images were unseen during QuPAINT training, and the agent adaptation uses synthetic conversations only.
\Cref{tab:workloads} summarizes the evaluation tasks.

\begin{table}[t]
\centering
\caption{Evaluation workloads. Characterization uses the same 24 images as interactive inspection.}
\label{tab:workloads}
\small
\renewcommand{\arraystretch}{1.12}
\begin{tabularx}{\columnwidth}{@{}>{\raggedright\arraybackslash}p{0.33\columnwidth}r>{\raggedright\arraybackslash}X@{}}
\toprule
Workload & Cases & Evaluation focus \\
\midrule
Record reasoning & 60 & Tool use and answer correctness \\
Positive-image inspection & 12 & Five successive requests \\
Negative-image inspection & 12 & Images without annotated monolayers \\
Characterization & 24 & Localization and original-model agreement \\
\bottomrule
\end{tabularx}
\end{table}

\subsection{Comparison Methods}
We compare OpenQlaw-v2 with a two-model workflow using Qwen3-VL-32B-Instruct as the orchestrator and QuPAINT-4B as the domain expert, and with unadapted QuPAINT as the agent.
All conditions use the same tools and evaluation requests. Record-based tasks also share identical observations.
Each condition receives the same explicit tool-call format and argument definitions.
The unadapted control tests whether domain specialization alone provides agent behavior, while the two-model control represents the original orchestration design.

\subsection{Evaluation Metrics}
We report task completion, correct tool execution and answer correctness, including appropriate refusals when required information is unavailable.
Strict task success requires every requested turn to complete, its tool calls and arguments to match the reference trace, and its answer to satisfy the task-specific checks.
Refusal-aware success additionally accepts a direct clarification without a tool call for requests with missing calibration, missing labels, mismatched image records or unavailable candidate enumeration.
The clarification must identify the missing information. An attempted unsupported operation does not qualify as a direct refusal.


A successful image workflow must complete all five requests using consistent candidate identities and measurements.
These checks use each condition's admitted observations and actual tool results. Regarding this, they do not require its candidate list to match the image annotations.
The negative sample images receive the initial inspection request only and are excluded from the five-request success denominator.
Their saved traces are reviewed separately for positive-count results, unavailable records and unfinished requests.
Monolayer localization is summarized by recall at an intersection-over-union (IoU) threshold of $0.25$ and mean best-match IoU over reference flakes, assigning zero when no usable prediction is available. These measures do not penalize extra predicted boxes.
We separately assess output agreement with the original QuPAINT checkpoint, response latency and GPU and host memory use.
We report 95\% Wilson intervals for positive-image workflow success, treating each image as one case.
The record-based episodes share six source scenes and task templates, which limits conclusions about unfamiliar requests.

\subsection{Implementation Details}
The adaptation curriculum contains 880 synthetic conversations, of which 700 conversations from 260 scenes provide 1,890 training targets.
The remaining 180 conversations form separate development partitions.
The training set contains 400 single-request and 300 multi-request conversations. Each target is one assistant response conditioned on the preceding context, as defined in \Cref{eq:training}.
The language adapter uses rank $16$, scale $32$ and dropout $0.05$.
It adapts the query, key, value and output projections of the language attention layers, with 11M trainable parameters.
The visual encoder, physics-aware components and original language weights remain frozen.
The final adaptation stage continues the development adapter for 256 steps using AdamW with a learning rate of $10^{-4}$.
The development adapter has already been trained on record operations, analysis-tool invocation, ranking and longer inspection histories. The 256 steps are its final continuation.
Experiments use one 40\,GB A100 GPU per condition. The 32B orchestrator additionally requires CPU offload.
Latency and memory comparisons therefore reflect these deployment configurations.
Resident and peak GPU values refer to allocated memory. Host memory is the cumulative peak resident set size of the process.
Each condition has one recorded deployment run, so the resource results do not estimate variability across repeated runs.
\section{Results}
\label{sec:results}

\begin{figure*}[!ht]
\centering
\includegraphics[width=\textwidth]{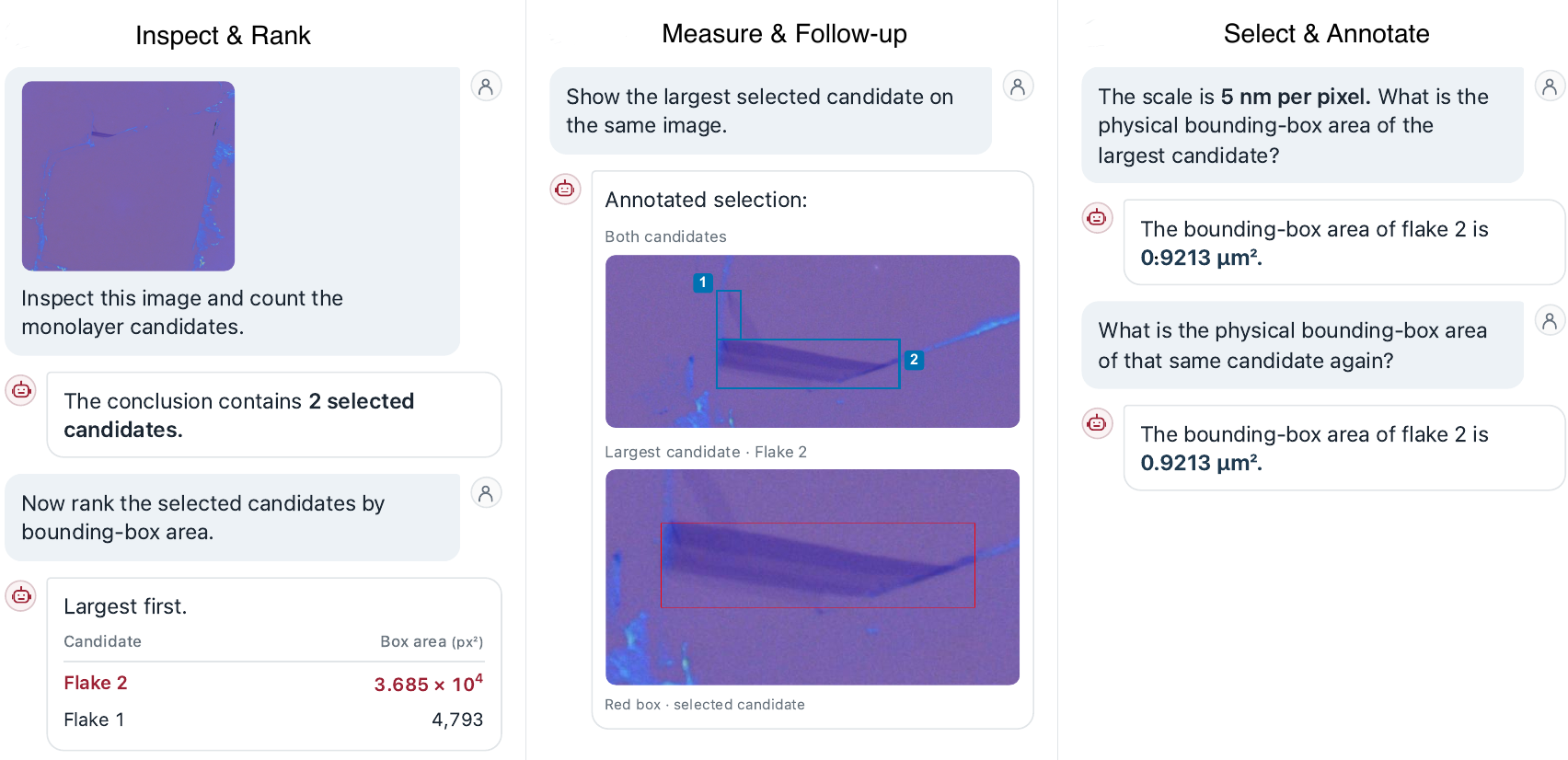}
\caption{\textbf{Interactive inspection with OpenQlaw-v2.} Recorded workflow shows both model-selected candidates, the largest-candidate annotation and successive measurement requests. The scale is supplied by the user.}
\label{fig:qualitative-interaction}
\end{figure*}

\subsection{Agent Performance}
\label{sec:mainresults}
\Cref{tab:workflowresults} compares OpenQlaw-v2 with the two-model workflow and unadapted QuPAINT.
OpenQlaw-v2 achieves $80.0\%$ strict task success and $90.0\%$ refusal-aware success. The unadapted model, in comparison, only yields $40.0\%$ for both measures.
The two-model workflow achieves $90.0\%$ on both measures.
Thus, adaptation substantially improves QuPAINT's ability to control the inspection tools, while the separate orchestrator retains an advantage under the strict reference-trace criterion.

\begin{table*}[t]
\centering
\caption{Agent performance on 60 record-based episodes and 12 positive-image workflows. Image-workflow success requires completing all five requests. Percentages are shown in parentheses.}
\label{tab:workflowresults}
\small
\renewcommand{\arraystretch}{1.12}
\begin{tabularx}{\textwidth}{@{}Xrrr@{}}
\toprule
Agent configuration & Strict Task Success & Refusal-Aware Success & Image Workflow Success \\
\midrule
Unadapted QuPAINT & 24/60 (40.0) & 24/60 (40.0) & 0/12 (0.0) \\
Qwen3-VL-32B + QuPAINT & 54/60 (90.0) & 54/60 (90.0) & 12/12 (100.0) \\
OpenQlaw-v2 & 48/60 (80.0) & 54/60 (90.0) & 10/12 (83.3) \\
\bottomrule
\end{tabularx}
\end{table*}

\noindent\textbf{Task Reasoning.}
For OpenQlaw-v2, six image-mismatch cases receive appropriate clarification requests instead of the failing reference operation.
These responses pass the refusal-aware check but not strict trace matching.
The remaining six failures concern missing layer labels. The agent acknowledges that labels are absent but proposes counting the full record instead of clearly declining the layer-specific request.
The two-model workflow's six failures concern missing calibration, where it proposes a scale not supplied by the user. The tool rejects this value before the model asks for calibration.

\subsection{Interactive Image Inspection}
\label{sec:imageworkflows}
OpenQlaw-v2 passes all five requested operations and answer checks on ten of 12 positive images ($83.3\%$).
The two-model workflow passes on all 12 images, while unadapted QuPAINT passes on none.
The adapted model correctly counts and ranks the admitted candidates on every positive image.
Its two unsuccessful workflows stop at annotation because the tool call omits the required candidate IDs.
The subsequent measurement requests are therefore not completed in these cases.
The 95\% Wilson intervals for complete-workflow success are $55.2$-$95.3\%$ for OpenQlaw-v2, $75.8$-$100.0\%$ for the two-model workflow and $0.0$-$24.2\%$ for unadapted QuPAINT.
These intervals reflect the small image cohort. 

\Cref{fig:qualitative-interaction} shows a recorded interaction in which OpenQlaw-v2 compares two candidates, annotates the largest and answers successive measurement requests.
Each adapted positive-image workflow invokes image analysis once and uses the retained observations for subsequent requests.
Five successful workflows reuse the preceding measurement in the final answer. The other five repeat the deterministic measurement.
These results demonstrate interactive tool use over retained observations.

\noindent\textbf{Negative Samples.}
Negative-images are images that do not contain any annotated monolayers. A separate review of the 12 negative-image traces finds a positive count returned by the scientific tool in one OpenQlaw-v2 case, compared with ten cases for each control.
These counts describe tool results over the records each agent selected, rather than a correct-rejection rate.
In ten OpenQlaw-v2 cases, the conclusion record is unavailable for counting because it has no enumerated candidates. The agent requests further information.
One additional case yields invalid analysis boxes and no usable record.
The two-model workflow also exhausts its request budget on one negative image.
Thus, declining an operation over an unavailable record should not be interpreted as establishing that the image contains no monolayer flakes.

\subsection{Computational Requirements}
\label{sec:resources}
\begin{figure*}[htbp]
\centering
\includegraphics[width=\textwidth]{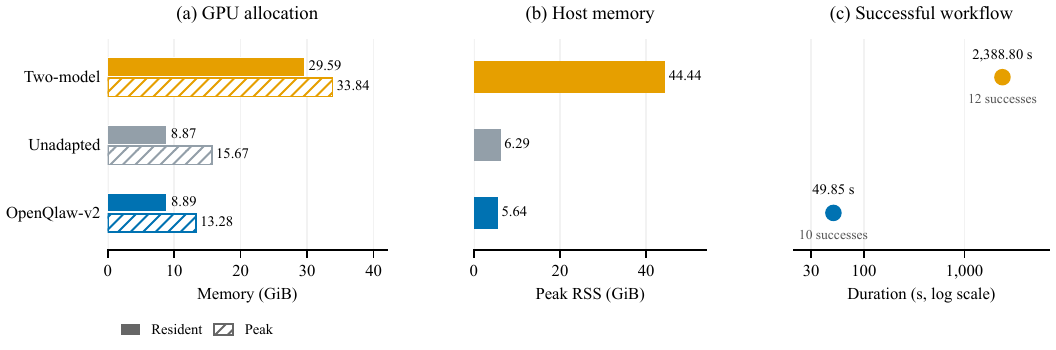}
\caption{\textbf{Resource comparison in the tested deployments.} \textbf{(a)} Resident and peak allocated GPU memory \textbf{(b)} peak host memory, including preceding development checks \textbf{(c)} median successful five-request workflow duration on a logarithmic axis. The two-model configuration uses CPU offload. Timing includes 12 successful images for the two-model workflow and ten for OpenQlaw-v2. The unadapted QuPAINT completes no full workflow and has no successful-workflow median.}
\label{fig:resources}
\end{figure*}

\Cref{tab:resources} and \Cref{fig:resources} compare the measured deployment requirements.
OpenQlaw-v2 uses one loaded checkpoint for both agent reasoning and material analysis, compared with two checkpoints in the original workflow.

\noindent\textbf{GPU and Host Memory.}
Resident GPU memory decreases from $29.59$ to $8.89\,\mathrm{GiB}$, a reduction of $70.0\%$.
Peak allocated GPU memory decreases from $33.84$ to $13.28\,\mathrm{GiB}$, a reduction of $60.7\%$.
The unadapted expert uses $8.87\,\mathrm{GiB}$ of resident GPU memory, so the adapted agent adds approximately $0.02\,\mathrm{GiB}$ to the measured resident allocation.
This shows that agent adaptation retains a memory footprint close to the existing expert while removing the separately hosted orchestrator.
Peak host memory is $5.64\,\mathrm{GiB}$ for OpenQlaw-v2 and $44.44\,\mathrm{GiB}$ for the two-model configuration.
The latter offloads part of the 32B orchestrator to the CPU. These host measurements are cumulative process peaks that include preceding development checks.

\begin{table}[t]
\centering
\caption{Deployment measurements. Memory is in GiB. Time is the median successful five-request workflow in seconds. The two-model reference uses CPU offload. Host memory includes development checks.}
\label{tab:resources}
\small
\setlength{\tabcolsep}{4pt}
\renewcommand{\arraystretch}{1.12}
\begin{tabularx}{\columnwidth}{@{}Xrrr@{}}
\toprule
Metric & Unadapted & Two-Model & v2 \\
\midrule
Checkpoints & 1 & 2 & 1 \\
Resident GPU & 8.87 & 29.59 & 8.89 \\
Peak GPU & 15.67 & 33.84 & 13.28 \\
Peak host & 6.29 & 44.44 & 5.64 \\
Workflow (s) & - & 2388.80 & 49.85 \\
\bottomrule
\end{tabularx}
\end{table}

\noindent\textbf{Workflow Duration and Performance Trade-Off.}
The median successful five-request workflow takes $49.85\,\mathrm{s}$ with OpenQlaw-v2 and $2{,}388.80\,\mathrm{s}$ with the two-model workflow.
The two-model configuration uses CPU offload, so these durations describe the tested deployments rather than an architecture-only speedup.
Successful-workflow medians also use different subsets: ten images for OpenQlaw-v2 and 12 for the two-model workflow.
A matched comparison restricted to the ten images completed by both conditions gives the same medians, $49.85$ and $2{,}388.80\,\mathrm{s}$, respectively.
This subset check removes the difference in image membership but still excludes unsuccessful workflows and retains the CPU-offload effect.
Together with \Cref{tab:workflowresults}, the measurements show a trade-off. The unified model greatly reduces serving requirements while completing slighlty fewer image workflows than the separately hosted orchestrator.
The comparison also changes the orchestrator size. The original workflow uses a 32B general-purpose model, while the unified system uses the existing 4B domain expert.
A smaller general-purpose orchestrator may offer a different trade-off and is not evaluated here.

\subsection{Preserving Material-Analysis Behavior}
\label{sec:characterization}
\Cref{tab:characterization} examines the original image-analysis path after adding agent functionality.
With the adapter disabled during characterization, OpenQlaw-v2 produces identical fixed-prompt output text to the original QuPAINT checkpoint on all 24 images.
Both conditions produce usable outputs on 22 images and identical localization scores.
This confirms preservation of the original characterization outputs on the evaluated images.

\begin{table}[t]
\centering
\caption{Characterization preservation on 24 unseen images. Localization is summarized by recall at IoU $\geq0.25$ and mean best-match IoU. Output agreement compares fixed-prompt text with the original checkpoint.}
\label{tab:characterization}
\small
\renewcommand{\arraystretch}{1.12}
\begin{tabularx}{\columnwidth}{@{}Xrr@{}}
\toprule
Metric & Original & OpenQlaw-v2 \\
\midrule
Usable outputs & 22/24 & 22/24 \\
Output agreement & Reference & 24/24 \\
\midrule
Recall at IoU $0.25$ (\%) & 47.1 & 47.1 \\
Mean best-match IoU & 0.290 & 0.290 \\
\bottomrule
\end{tabularx}
\end{table}

Recall at IoU $\geq0.25$ is $47.1\%$ (eight of 17 reference flakes), and mean best-match IoU is $0.290$ across all reference flakes, assigning zero when no usable prediction is available.
These measures describe spatial coverage at a relaxed overlap threshold and the overlap of each reference flake with its best predicted box.
The extension therefore adds agent functionality without changing the tested characterization outputs, while retaining the underlying model's localization limitations.

\subsection{Implications for Manufacturing \& Materials Inspection}
For fabrication planning, the workflow connects a detailed expert-level microscopy optical observation to a candidate selection, its location and its calibrated bounding-box dimensions.
Keeping the image, candidate identity and scale together allows laboratories to refine requests without manually transferring the preceding result between operations.
The interaction in \Cref{fig:qualitative-interaction} illustrates these steps, while \Cref{tab:workflowresults} and \Cref{fig:resources} show their measured performance and resource requirements.

The main benefit of consolidation is that this interaction can be served by the existing domain checkpoint with a small language adapter.
The tools still perform the numerical operations, and the retained records still provide the inspection context.
The adaptation changes which model coordinates these components.
This distinction matters when extending a laboratory workflow. Adding an agent interface does not require replacing the existing characterization model or its measurement tools.

For our experiments, we focus on the interactive analysis of saved microscopy images.
We do not measure fabricated-device yield, instrument control or production throughput.
\section{Conclusion}
\label{sec:conclusion}
In this work, we have introduced \method, which adapts QuPAINT to serve as the central reasoning agent for interactive inspection of 2D materials.
We have shown how to retain OpenQlaw's scientific tools and image-specific records while supporting task reasoning and material analysis within one loaded checkpoint.
Our approach has achieved $90.0\%$ refusal-aware record-task success and completed ten of 12 image workflows, compared to $90.0\%$ and 12 of 12 for the two-model reference.
Resident GPU memory is approximately $70\%$ lower in the measured deployment, while original characterization outputs have been preserved on all 24 tested images.
These findings demonstrate that a domain-specific model can coordinate interactive inspection with substantially lower deployment requirements, although some workflow reliability is sacrificed.

\noindent\textbf{Limitations.}
Constructed tool-use conversations may not capture the variety of requests encountered during laboratory work.
Longer interactions and requests outside the training task families require further evaluation.
Our experiments focus on saved microscopy images and do not assess fabricated-device yield, instrument control, or production throughput.

\section*{Acknowledgments}
This work is partly supported by MonArk NSF Quantum Foundry (DMR-1906383), NSF PCL-Test Bed AURORA (2607535) and  NSF Quantum Award (2444042). It acknowledges the Arkansas High-Performance Computing Center for providing GPUs.

\bibliographystyle{asmeconf}
\bibliography{references}

\end{document}